\documentclass[sigconf, preprint]{acmart}

\renewcommand\footnotetextcopyrightpermission[1]{} 

\usepackage{multirow}

\begin{document}

\title{EviStress: Evidence Stress Learning for Generalizable Face Forgery Detection with Textual Explanations}

\newcommand{\myaffiliation}{%
  \affiliation{%
    \institution{School of Cyber Engineering, Xidian University}
    \city{Xi'an}
    \country{China}
  }%
}
\author{Jiuyao Jing}
\myaffiliation
\email{25151110112@stu.xidian.edu.cn}

\author{Xuanyu Yi}
\myaffiliation
\email{25189100096@stu.xidian.edu.cn}

\author{Yu Zheng}
\myaffiliation
\email{yzheng@xidian.edu.cn}

\author{Chunlei Peng}
\correspondingauthor
\myaffiliation
\email{clpeng@xidian.edu.cn}

\renewcommand{\shortauthors}{Trovato et al.}

\begin{abstract}
The rapid spread of face forgeries on the Web challenges the credibility of visual content. In practical Web environments, manipulated media originate from evolving generation pipelines and undergo diverse processing during dissemination, making reliable authenticity assessment under unseen manipulations and visual changes a critical challenge. We observe that a standard LoRA-adapted CLIP detector exhibits uneven patch-level decision dependence, concentrating a disproportionate share of class-specific relevance on a small fraction of visual patches. To improve robustness to variations in such highly dependent evidence, we propose EviStress, a two-stage framework for generalizable face forgery detection with textual explanations. In the first stage, Evidence-Targeted Patch Perturbation (ETPP) estimates patch dependence from global, relational, and semantic cues and assigns positively correlated perturbation probabilities, explicitly stress-testing highly dependent visual evidence during training. We further introduce Discriminative Variation Preservation (DVP), which constrains the residual between clean and perturbed representations to retain authenticity-discriminative information. In the second stage, Intrinsic Evidence Instruction Tuning (IEIT) transfers the adapted CLIP encoder into LLaVA as its vision tower and performs parameter-efficient multimodal instruction tuning, enabling authenticity prediction and textual explanation directly from the learned forgery-sensitive representation without an external detector. Experiments demonstrate improved cross-dataset and cross-method generalization and corruption robustness, with the multimodal model providing authenticity predictions and human-readable explanations.


\end{abstract}



\maketitle

\pagestyle{empty}

\section{Introduction}
\label{sec:intro}

The rapid progress of generative models has made realistic face manipulation increasingly accessible and widely distributable through online platforms, posing growing threats to Web content credibility and public trust. Media encountered online may originate from previously unseen generation pipelines and further undergo heterogeneous resizing, compression, transcoding, and re-encoding during dissemination. Consequently, reliable authenticity assessment must account not only for unseen manipulation methods, but also for changes in visual evidence induced by online content processing. This work addresses the content-authenticity aspect of Web security, where generalization beyond the training distribution is a central requirement for practical face forgery detection~\cite{deepfakebench,ucf,lsda,sbi,prodet}.


\begin{figure}[t]
    \centering
    \includegraphics[width=1\linewidth]{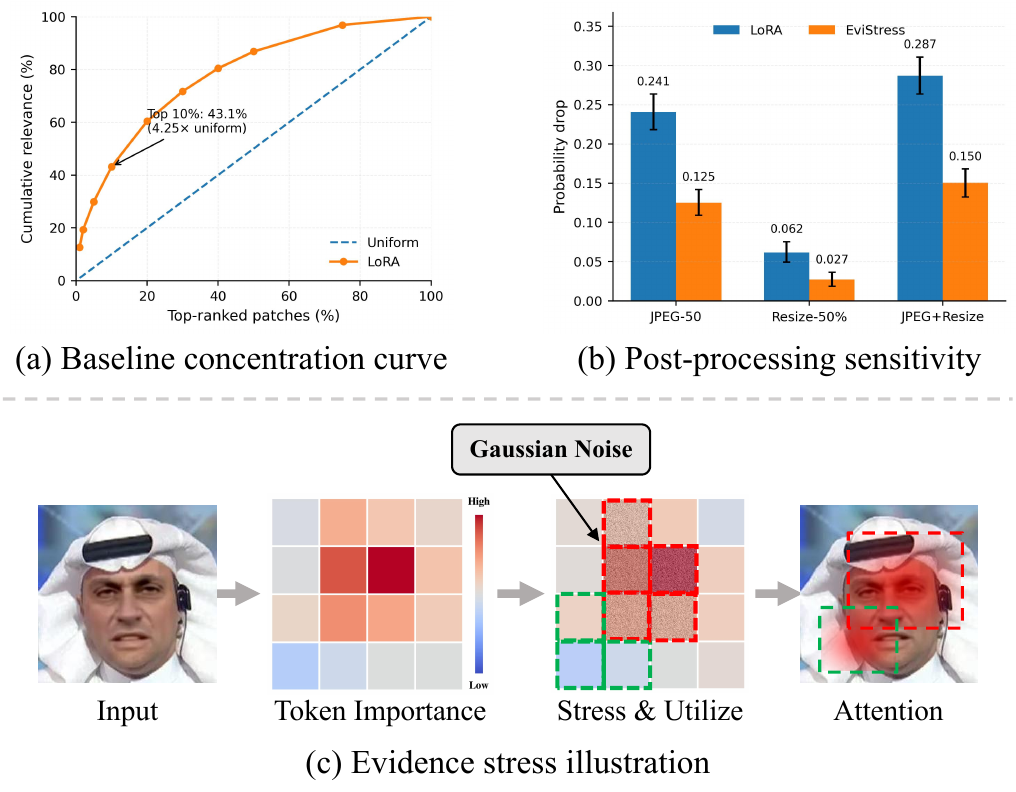}
    \caption{Motivation and evidence stress in EviStress. (a) LoRA-adapted CLIP concentrates patch-level relevance: top 10\% patches account for 43.1\% of total relevance vs. 10.2\% under uniform allocation. (b) The EviStress visual detector shows smaller probability drops under evaluated JPEG compression, resizing, and combined transformations. (c) Evidence stress is that training preferentially applies Gaussian perturbations to highly dependent patch tokens.}
    \label{fig:intro}
    \Description{
    Three panels show patch-level relevance concentration in a LoRA baseline, robustness under Web-style transformations, and the evidence-targeted patch perturbation mechanism.
    }
\end{figure}

Recent studies increasingly build generalizable detectors upon vision foundation models and vision--language models, particularly CLIP~\cite{radford2021learning}. Its large-scale pretraining provides transferable visual and semantic representations, while subsequent methods improve forgery modeling through parameter-efficient adaptation, local token learning, token reorganization, and pretrained-prior preservation~\cite{yan2024orthogonal,forada,udd,qtfp,fia-usa}. However, improving access to local evidence does not necessarily make the resulting decision robust to changes in that evidence. To examine how a source-trained detector actually uses local visual cues, we quantitatively analyze class-specific patch relevance in a standard LoRA-adapted CLIP detector. As shown in Figure~\ref{fig:intro}(a), its decision relevance is markedly non-uniform: the top 10\% patches account for 43.1\% of the total class-margin relevance, approximately 4.25 times the mass expected under uniform allocation. Moreover, intervention on the most supportive patches causes greater prediction degradation than removing the same number of randomly selected patches, indicating that these high-relevance regions influence the detector’s authenticity predictions rather than merely being visually salient. These observations motivate a more direct question: instead of forcing the detector to distribute its relevance uniformly, can we improve generalization by repeatedly stress-testing the visual evidence on which its current decision depends most strongly?

Beyond a binary label, textual explanations provide a human-readable complement to online content authenticity assessment. Multimodal large language models offer a way to generate authenticity predictions and explanations within a single framework. Some approaches retain specialized vision detectors for authenticity prediction and employ the language model mainly as an interpreter or feature aggregation module~\cite{guo2025rethinking,dd-vqa,huang2024ffaa,peng2025mllm}. Although this design benefits from established detectors, its core detection capability remains external to the MLLM. Other studies directly fine-tune MLLMs with curated forgery image--text data to jointly produce predictions and explanations~\cite{fftg,huang2025sida,xu2025fakeshield,zhou2025aigi}. However, generic MLLM vision encoders are primarily pretrained for semantic understanding and are not explicitly optimized for subtle, low-level facial manipulation traces~\cite{ren2025can,tariq2025llms}. Consequently, multimodal instruction tuning alone may yield plausible rationales without ensuring that the underlying authenticity prediction is built upon a forgery-sensitive visual representation extracted from the input image.

To address these issues, we propose \textbf{EviStress}, a two-stage framework for generalizable face forgery detection with textual explanations. In the first stage, we introduce \textbf{Evidence-Targeted Patch Perturbation} (ETPP) to explicitly stress-test highly dependent visual evidence. ETPP estimates patch dependence through global, relational, and semantic cues and assigns perturbation probabilities that are positively correlated with the resulting dependence scores. Instead of uniformly perturbing visual tokens, ETPP preferentially applies controlled stochastic modulation to patches with stronger estimated dependence, repeatedly exposing the detector to variations of influential local evidence during training, as illustrated in Figure~\ref{fig:intro}(c). Since targeted intervention may also introduce class-irrelevant feature variations, we further propose \textbf{Discriminative Variation Preservation} (DVP), which constrains the residual between clean and perturbed representations to retain authenticity-discriminative information. As shown in Figure~\ref{fig:intro}(b), the adapted visual detector exhibits smaller probability drops under the evaluated post-processing transformations. In the second stage, we present \textbf{Intrinsic Evidence Instruction Tuning} (IEIT) for self-contained explainability. IEIT integrates the adapted CLIP encoder into LLaVA as its vision tower, freezes the visual encoder, and optimizes the multimodal projector and the LLM through LoRA, thereby transferring the learned forgery-sensitive representation to multimodal instruction tuning.

Our main contributions are summarized as follows:

\begin{itemize}
    \item We propose EviStress, a two-stage evidence stress learning framework for generalizable face forgery detection with textual explanations, aimed at supporting Web content authenticity assessment.

    \item We introduce \textbf{ETPP}, which preferentially stress-tests highly dependent visual evidence during training, together with \textbf{DVP}, which preserves authenticity-discriminative information in the induced representation variation. We further develop \textbf{IEIT} to transfer the adapted visual encoder to an MLLM without relying on an external detector.

    \item We quantitatively characterize concentrated patch-level decision dependence and show that evidence stress learning improves cross-dataset and cross-method generalization and average robustness under controlled image corruptions. We further assess textual explanation quality through automatic scoring and a blinded human ranking study against the evaluated open-source MLLMs.
\end{itemize}

\section{Related Work}
\label{sec:related}

\subsection{Face Forgery Detection}

Face forgery detection seeks to identify manipulated facial content from subtle visual and temporal inconsistencies. Early approaches mainly exploit low-level artifacts, including color mismatch, abnormal head geometry, and blending boundaries~\cite{li2018exposing,yang2019exposing,wu2020sstnet}. However, such cues are often tied to specific manipulation pipelines and generalize poorly beyond the training distribution. To improve robustness, subsequent studies explore frequency-aware representations~\cite{liu2021spatial,freqdebias}, pseudo-forgery generation and data augmentation~\cite{sbi,lsda}, disentangled feature learning~\cite{hong2024contrastive,li2025critical}, and temporal or audio-visual consistency~\cite{oorloff2024avff,vb-sta}. Despite improved performance, these methods remain vulnerable to manipulation mechanisms and distribution shifts that are absent from the training data.

More recently, vision foundation models and vision--language models, particularly CLIP~\cite{radford2021learning}, have provided transferable visual and semantic priors for forgery detection. Effort~\cite{yan2024orthogonal} decomposes pretrained representations into orthogonal subspaces, adapting forgery-related components while preserving general visual priors. ForAda~\cite{forada} introduces lightweight task-specific modules to inject forensic traces into CLIP features. UDD~\cite{udd} shuffles and mixes patch tokens to reduce positional and content biases. QTFP~\cite{qtfp} introduces learnable query tokens to aggregate informative local evidence instead of relying solely on the global token. FIA-USA~\cite{fia-usa} constructs pseudo-forgeries by simulating transferable facial inconsistencies and up-sampling artifacts. These methods primarily improve evidence extraction, aggregation, or representation adaptation. In contrast, the robustness of decision-critical local evidence to visual variation is rarely an explicit training objective, leaving detectors vulnerable to key-evidence shifts in unseen distributions.

\begin{figure*}[ht]
    \centering
    \includegraphics[width=1\linewidth]{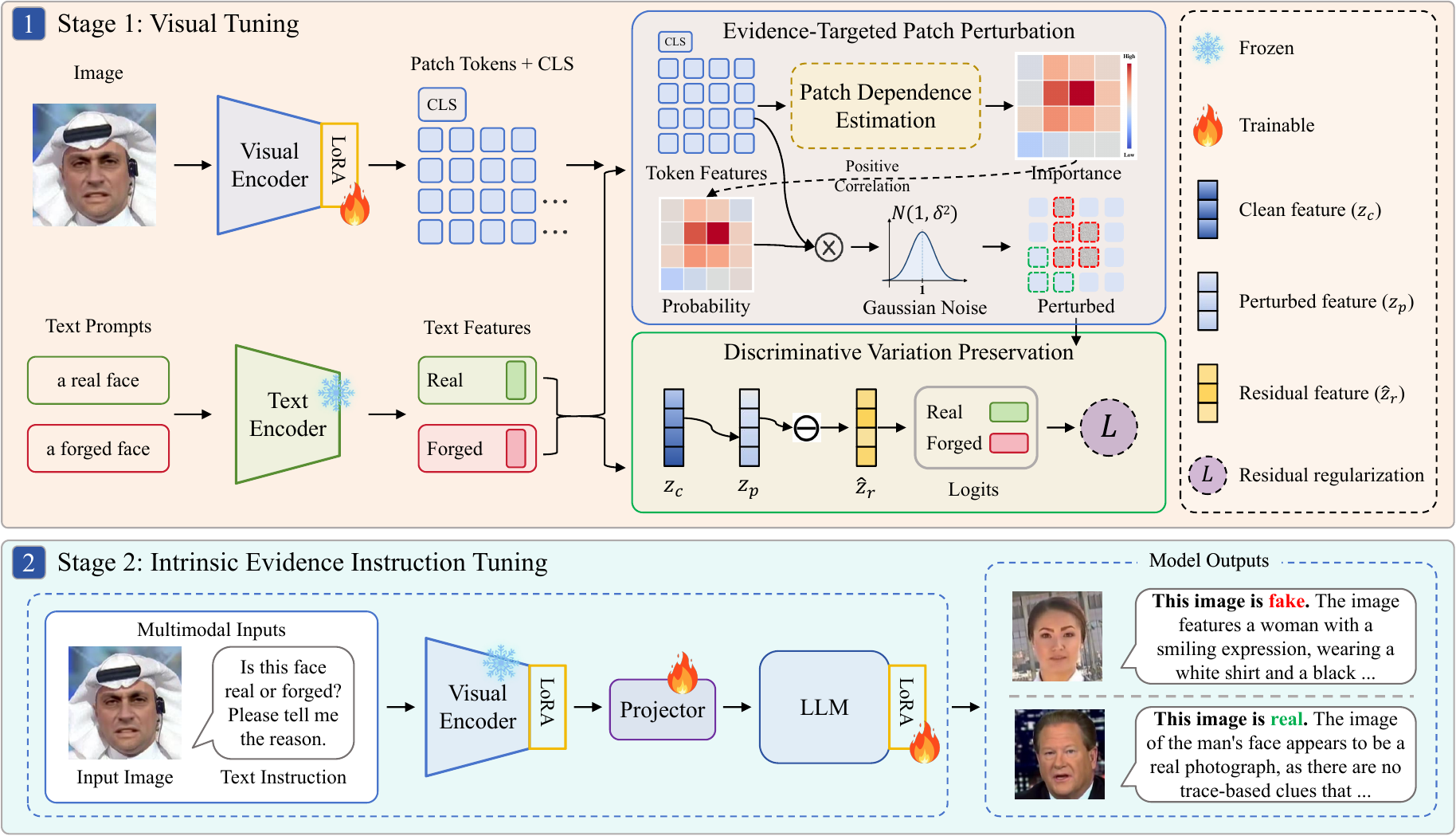}
    \caption{Overview of the EviStress framework. In Stage 1, Evidence-Targeted Patch Perturbation (ETPP) estimates patch dependence and stress-tests highly dependent visual evidence, while Discriminative Variation Preservation (DVP) preserves authenticity-discriminative information in the induced variation. In Stage 2, the adapted visual encoder is integrated into a multimodal large language model for joint authenticity prediction and textual explanation.}
    \label{fig:model}
    \Description{A woman and a girl in white dresses sit in an open car.}
\end{figure*}

\subsection{Explainable Forgery Detection}

Explainable forgery detection extends authenticity classification with human-readable descriptions of the supporting evidence. One line of work retains specialized visual detectors or fusion modules for the core decision and uses language models mainly for interpretation. DD-VQA~\cite{dd-vqa} formulates detection as visual question answering with common-sense descriptions of facial abnormalities. M2F2-Det~\cite{guo2025rethinking} combines CLIP-based authenticity prediction with language-based interpretation, while FFAA~\cite{huang2024ffaa} and VLF-FFD~\cite{peng2025mllm} aggregate visual and textual information for open-world detection and explanation. These approaches improve interpretability, but the principal detection capability remains partly outside the MLLM.

Another line directly fine-tunes MLLMs with forgery-oriented image--text data. FFTG~\cite{fftg} constructs fine-grained descriptions using manipulation and region annotations. SIDA~\cite{huang2025sida} jointly performs detection, localization, and explanation for social-media images. FakeShield~\cite{xu2025fakeshield} unifies forgery analysis and localization, while AIGI-Holmes~\cite{zhou2025aigi} combines visual-expert adaptation and multimodal instruction tuning for explainable generated-image detection. However, general-purpose MLLM vision encoders remain insufficiently sensitive to subtle facial manipulation traces~\cite{ren2025can,tariq2025llms}. These developments motivate learning a visual representation that is not only forgery-sensitive but also robust to variations in influential local evidence. We address this objective through evidence stress learning before multimodal instruction tuning, integrating the adapted encoder into the MLLM for joint authenticity prediction and textual explanation.

\section{Methodology}
\label{sec:method}

\subsection{Evidence-Targeted Patch Perturbation}
\label{sec:etpp}

Our quantitative analysis shows that a source-trained detector can exhibit markedly uneven patch-level decision dependence, with a limited fraction of visual patches carrying a disproportionate share of class-specific relevance. Intervention analysis further confirms that highly ranked supportive patches exert substantially greater influence on the authenticity decision than randomly selected patches. We therefore propose \textbf{Evidence-Targeted Patch Perturbation} (ETPP), which preferentially stress-tests the visual patches on which the detector exhibits stronger estimated dependence, as illustrated in Figure~\ref{fig:model}.

Given an input face image \(I\), the visual encoder \(\mathcal{V}_{\theta}\), consisting of \(L\) Transformer blocks, produces one global token and \(N\) patch tokens at each block. We denote the token sequence after the \(l\)-th block as
\begin{equation}
    \mathbf{X}^{(l)}
    =
    [\mathbf{x}_{\mathrm{cls}}^{(l)},
    \mathbf{x}_{1}^{(l)},\ldots,\mathbf{x}_{N}^{(l)}],
    \qquad
    \mathbf{x}_{i}^{(l)}\in\mathbb{R}^{d}.
\end{equation}
ETPP is applied to the first \(K\) Transformer blocks, i.e.,
\(l\in\{1,\ldots,K\}\), where \(K\leq L\) controls the perturbation depth.
Patch Dependence Estimation assigns each patch a normalized dependence score
\(s_i^{(l)}\in[0,1]\), which is linearly mapped to the perturbation probability
as \(p_i^{(l)}=p_{\min}+s_i^{(l)}(p_{\max}-p_{\min})\), with
\(p_{\min}=0.45\) and \(p_{\max}=0.95\). Hence, patches with stronger estimated
dependence are perturbed more frequently.

For each patch token in the selected blocks, we sample a binary indicator \(m_i^{(l)}\sim\operatorname{Bernoulli}(p_i^{(l)})\) and a multiplicative factor \(\epsilon_i^{(l)}\sim\mathcal{N}(1,\delta^2)\). The perturbed token is defined as:
\begin{equation}
    \widetilde{\mathbf{x}}_{i}^{(l)}
    =
    \bigl[(1-m_i^{(l)})+
    m_i^{(l)}\epsilon_i^{(l)}\bigr]
    \mathbf{x}_{i}^{(l)},
    \qquad l\leq K.
\end{equation}
The global token remains unchanged, and no perturbation is applied to blocks with \(l>K\). Since the Gaussian factor is centered at one, the intervention introduces controlled feature variation without deterministically suppressing the selected evidence.

The clean and perturbed token sequences are then processed by the remaining visual aggregation and projection operations, producing the clean representation \(\mathbf{z}_{c}\) and the evidence-stressed representation \(\mathbf{z}_{p}\), respectively. Rather than enforcing invariance between them, we retain the variation induced by targeted evidence stress and regulate its discriminative content through DVP.

\begin{figure}[t]
    \centering
    \includegraphics[width=1\columnwidth]{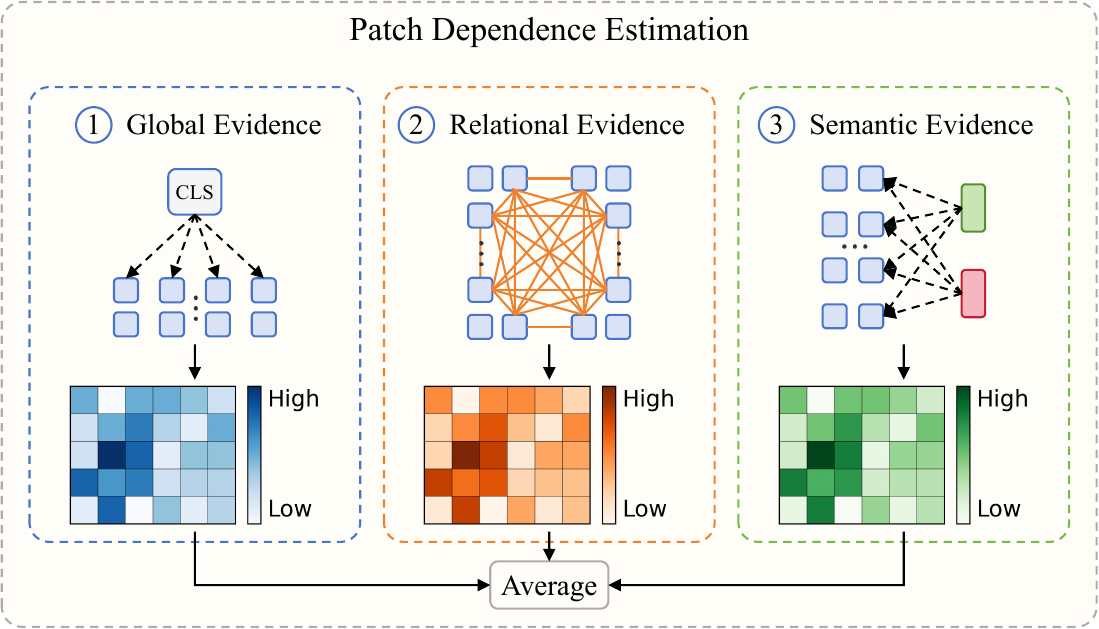}
    \caption{Multi-cue patch dependence estimation in ETPP. Patch dependence, derived from global, relational, and semantic cues, determines tokens' perturbation probability.}
    \label{fig:pde}
    \Description{This figure illustrates the approach to patch dependence estimation.}
\end{figure}

\subsection{Patch Dependence Estimation}
\label{sec:pde}

ETPP requires an operational estimate of patch dependence that can be computed during visual encoding. A single criterion may be incomplete: a patch can be strongly associated with the global representation without carrying clear authenticity semantics, while a locally discriminative patch may not be fully characterized by its relationship with the global token. We therefore estimate patch dependence from three complementary cues---global, relational, and semantic---as illustrated in Figure~\ref{fig:pde}.

Let \(\operatorname{sim}(\cdot,\cdot)\) denote cosine similarity. For the \(i\)-th patch token, the three cue scores are defined as:
\begin{equation}
\begin{aligned}
    u_i^{g}
    &=
    \operatorname{sim}
    (\mathbf{x}_{i},\mathbf{x}_{\mathrm{cls}}),\\
    u_i^{r}
    &=
    \frac{1}{N-1}
    \sum_{\substack{j=1\\j\neq i}}^{N}
    \operatorname{sim}
    (\mathbf{x}_{i},\mathbf{x}_{j}),\\
    u_i^{s}
    &=
    \left|
        \operatorname{sim}(\mathbf{x}_{i},\mathbf{t}_{f})
        -
        \operatorname{sim}(\mathbf{x}_{i},\mathbf{t}_{r})
    \right|,
\end{aligned}
\end{equation}
where \(\mathbf{t}_{r}\) and \(\mathbf{t}_{f}\) are the textual prototypes obtained from text prompts using the frozen text encoder.

The global cue measures the alignment between a local patch and the image-level representation. The relational cue characterizes its average association with the remaining visual tokens, capturing inter-patch structural consistency. The semantic cue measures its differential similarity to the real and forged textual prototypes, introducing authenticity-related semantics into the dependence estimate.

Because the three scores have different numerical ranges, each cue is independently normalized across the patch tokens. The final dependence score is obtained by equal-weight aggregation:
\begin{equation}
    s_i
    =
    \frac{1}{3}
    \left(
        \widehat{u}_{i}^{g}
        +
        \widehat{u}_{i}^{r}
        +
        \widehat{u}_{i}^{s}
    \right),
\end{equation}
where \(\widehat{u}_{i}^{g}\), \(\widehat{u}_{i}^{r}\), and \(\widehat{u}_{i}^{s}\) denote the normalized cue scores. Their combination provides a multi-cue training-time proxy for patch dependence by jointly considering global alignment, inter-patch association, and authenticity-related semantics. The resulting score determines the patch-wise perturbation probabilities used by ETPP.

\subsection{Discriminative Variation Preservation}
\label{sec:dvp}

ETPP deliberately introduces controlled variation into highly dependent visual evidence. However, not every perturbation-induced change is relevant to face authenticity, and excessive variation may distort useful forensic information. Directly minimizing the discrepancy between the clean and perturbed representations would instead suppress the intended evidence stress. We therefore propose \textbf{Discriminative Variation Preservation} (DVP), which constrains the perturbation-induced variation to retain real-versus-forged discriminability.

Given the clean representation \(\mathbf{z}_{c}\) and the perturbed representation \(\mathbf{z}_{p}\), we construct a normalized residual:
\begin{equation}
    \mathbf{z}_{r}
    =
    \mathbf{z}_{c}-\mathbf{z}_{p},
    \qquad
    \widehat{\mathbf{z}}_{r}
    =
    \frac{\mathbf{z}_{r}}
    {\lVert\mathbf{z}_{r}\rVert_{2}+\varepsilon},
\end{equation}
where \(\varepsilon\) is a small constant for numerical stability. This residual represents the change produced when highly dependent visual evidence is stressed.

DVP aligns the residual with the real and forged textual prototypes. Its class logits and supervised objective are defined as:
\begin{equation}
    \ell^{r}_{k}
    =
    \frac{
        \widehat{\mathbf{z}}_{r}^{\top}\mathbf{t}_{k}
    }{\tau},
    \qquad
    \mathcal{L}_{\mathrm{dvp}}
    =
    -\log
    \frac{
        \exp(\ell^{r}_{y})
    }{
        \sum_{k\in\{r,f\}}\exp(\ell^{r}_{k})
    },
\end{equation}
where \(k\in\{r,f\}\), \(y\) is the authenticity label, and \(\tau\) is the temperature parameter. This objective reduces the predictive uncertainty of the residual and prevents it from degenerating into arbitrary noise.

The clean representation is simultaneously optimized using the standard CLIP-based binary classification loss \(\mathcal{L}_{\mathrm{cls}}\). The complete first-stage objective is:
\begin{equation}
    \mathcal{L}_{\mathrm{stage1}}
    =
    \mathcal{L}_{\mathrm{cls}}
    +
    \lambda_{\mathrm{dvp}}
    \mathcal{L}_{\mathrm{dvp}},
\end{equation}
where \(\lambda_{\mathrm{dvp}}\) controls the strength of residual regularization. ETPP and DVP therefore play complementary roles, with ETPP introducing targeted variation into highly dependent visual evidence while DVP preserves authenticity-related structure within the resulting representation change.

\subsection{Intrinsic Evidence Instruction Tuning}
\label{sec:ieit}

After the first-stage evidence stress learning, we transfer the adapted visual encoder to a multimodal large language model for joint authenticity prediction and textual explanation generation. We refer to this procedure as \textbf{Intrinsic Evidence Instruction Tuning} (IEIT), because the MLLM is directly conditioned on the forgery-sensitive visual representation learned in Stage 1 rather than relying on an external detector.

The LoRA-adapted CLIP encoder learned in the first stage replaces the original LLaVA vision tower. Given an image and an instruction, the multimodal input is constructed as:
\begin{equation}
    \mathbf{H}_{i}
    =
    \left[
        \mathcal{P}_{\phi}
        \bigl(\mathcal{V}_{\theta_v}(I_i)\bigr);
        \mathcal{E}(\mathbf{q}_i)
    \right],
\end{equation}
where \(\mathcal{P}_{\phi}\) is the multimodal projector and \(\mathcal{E}(\cdot)\) denotes the language embedding operation. The projector maps visual tokens into the language-model embedding space so that they can be jointly processed with the textual instruction.

\begin{table}[t]
    \centering
    \caption{\textbf{Frame-level} cross-dataset AUC (\%) on CDF, DFD, and DFDC. The symbols {\dag} indicates that results are taken from~\cite{forada}. The best results are highlighted in bold.}
    \label{tab:traditional_frame}
    \begin{tabular}{lccccc}
        \toprule
        Method & Venue & CDF & DFD & DFDC & Avg. \\
        \midrule
        RECCE~\cite{recce}~{\dag}      & CVPR'22    & 73.2 & 81.2 & 71.3 & 75.2 \\
        SBI~\cite{sbi}~{\dag}        & CVPR'22    & 81.3 & 77.4 & --   & --   \\
        DCL~\cite{dcl}               & AAAI'22    & 82.3 & 91.7 & 76.7 & 83.6 \\
        UCF~\cite{ucf}~{\dag}        & ICCV'23    & 75.3 & 80.7 & 71.9 & 76.0 \\
        CFM~\cite{cfm}               & TIFS'24    & 82.8 & 91.5 & --   & --   \\
        LSDA~\cite{lsda}              & CVPR'24    & 83.0 & 88.0 & 73.6 & 81.5 \\
        ProDet~\cite{prodet}            & NeurIPS'24 & 84.5 & --   & 72.4 & --   \\
        DiffusionFake~\cite{diffusionfake}     & NeurIPS'24 & 80.5 & 90.4 & --   & --   \\
        UDD~\cite{udd}               & AAAI'25    & 86.9 & 91.0 & 75.8 & 84.6 \\
        ForAda~\cite{forada}~{\dag}     & CVPR'25    & 90.0 & 93.3 & 84.3 & 89.2 \\
        FreqDebias~\cite{freqdebias}        & CVPR'25    & 83.6 & 86.8 & 74.1 & 81.5 \\
        FFTG~\cite{fftg}              & CVPR'25    & 83.2 & \textbf{94.8} & --   & --   \\
        AdvOU~\cite{advou}             & ICCV'25    & 82.8 & 82.0 & 75.3 & 80.0 \\
        X$^{2}$DFD-7B~\cite{x2dfd}     & NeurIPS'25 & 90.4 & 92.3 & 83.7 & 88.8 \\
        X$^{2}$DFD-13B~\cite{x2dfd}    & NeurIPS'25 & \textbf{91.3} & 92.5 & 83.4 & 89.1 \\
        FIA-USA~\cite{fia-usa}           & NeurIPS'25 & 86.7 & 82.1 & --   & -- \\
        ResProto-FD~\cite{resproto}       & AAAI'26    & 83.8 & 91.4 & 79.5 & 84.9 \\
        Agent4FF~\cite{agent4faceforgery} & CVPR'26    & 87.1 & 93.3 & --   & --   \\
        SRI-Net~\cite{sri-net}           & ICLR'26    & 87.5 & 82.1 & --   & --   \\
        \midrule
        Ours (Stage 1)    & - & \textbf{91.3} & 93.7
                          & \textbf{85.1} & \textbf{90.0} \\
        Ours (Stage 2)    & - & 87.2 & 94.1 & 82.6 & 88.0 \\
        \bottomrule
    \end{tabular}
\end{table}

\paragraph{Instruction Tuning.}
The MLLM autoregressively generates the authenticity statement and its explanation. The training objective is:
\begin{equation}
    \mathcal{L}_{\mathrm{ieit}}
    =
    -\frac{1}{M}
    \sum_{i=1}^{M}
    \sum_{t=1}^{T_i}
    \log
    p_{\Theta}
    \left(
        a_{i,t}
        \mid
        a_{i,<t},
        \mathbf{q}_i,
        I_i
    \right),
\end{equation}
where \(a_{i,t}\) is the \(t\)-th target response token. The loss is computed only over the assistant response, while the image and instruction tokens serve as conditioning context.

During IEIT, the adapted visual encoder is frozen to preserve the forgery-sensitive representation learned through ETPP and DVP. At inference, the MLLM directly produces a unified response containing the authenticity prediction and its textual explanation.

\section{Experiments}
\label{sec:experiment}

\begin{table}[t]
    \centering
    \caption{\textbf{Video-level} cross-dataset AUC (\%) on CDF, DFD, and DFDC. The symbol~{\dag} indicates that results are taken from~\cite{qtfp}. The best results are highlighted in bold.}
    \label{tab:traditional_video}
    \begin{tabular}{lccccc}
        \toprule
        Method & Venue & CDF & DFD & DFDC & Avg. \\
        \midrule
        IID~\cite{iid}~{\dag}        & CVPR'23    & 83.8 & 93.9 & 70.0 & 82.6 \\
        UCF~\cite{ucf}~{\dag}        & ICCV'23    & 83.7 & 86.7 & 74.2 & 81.5 \\
        CFM~\cite{cfm}               & TIFS'24    & 89.7 & 95.2 & --   & --   \\
        LSDA~\cite{lsda}~{\dag}       & CVPR'24    & 87.5 & 88.1 & 77.0 & 84.2 \\
        CDFA~\cite{cdfa}~{\dag}       & ECCV'24    & 93.8 & 95.4 & 83.0 & 90.7 \\
        ProDet~\cite{prodet}            & NeurIPS'24 & 92.5 & --   & 77.0 & --   \\
        UDD~\cite{udd}               & AAAI'25    & 93.1 & 95.5 & 81.2 & 89.9 \\
        FCG~\cite{fcg}               & CVPR'25    & 95.0 & --   & 81.8 & --   \\
        VB-StA~\cite{vb-sta}            & CVPR'25    & 94.7 & 96.5 & 84.3 & 91.8 \\
        AdvOU~\cite{advou}             & ICCV'25    & 93.9 & 95.4 & 79.1 & 89.5 \\
        DeepShield~\cite{deepshield}        & ICCV'25    & 92.2 & 96.1 & 82.8 & 90.4 \\
        FakeRadar~\cite{fakeradar}         & ICCV'25    & 91.7 & 96.2 & 84.1 & 90.7 \\
        Effort~\cite{effort}            & ICML'25    & 95.6 & 96.5 & 84.3 & 92.1 \\
        X$^{2}$DFD-7B~\cite{x2dfd}     & NeurIPS'25 & 95.4 & 95.8 & 86.0 & 92.4 \\
        X$^{2}$DFD-13B~\cite{x2dfd}    & NeurIPS'25 & 95.7 & 96.1 & 85.7 & 92.5 \\
        FIA-USA~\cite{fia-usa}           & NeurIPS'25 & 94.1 & --   & 73.2 & --   \\
        ResProto-FD~\cite{resproto}       & AAAI'26    & 91.5 & 95.8 & 81.9 & 89.7 \\
        QTFP~\cite{qtfp}~{\dag}       & CVPR'26    & 96.0 & \textbf{97.5} & 86.9 & 93.5 \\
        SRI-Net~\cite{sri-net}           & ICLR'26    & 95.5 & 93.1 & --   & --   \\
        \midrule
        Ours (Stage 1)    & - & \textbf{97.8} & 96.8
                          & \textbf{88.2} & \textbf{94.2} \\
        Ours (Stage 2)    & - & 94.8 & \textbf{97.5} & 84.9 & 92.4 \\
        \bottomrule
    \end{tabular}
\end{table}

\subsection{Datasets}
\label{subsec:datasets}

For generalization evaluation, we follow prior work~\cite{deepfakebench,x2dfd,resproto} and train on the c23 version of FaceForensics++ (FF++)~\cite{rossler2019faceforensics++}. Celeb-DF-v2 (CDF)~\cite{cdf}, DeepFakeDetection (DFD)~\cite{dfd}, and the DeepFake Detection Challenge (DFDC)~\cite{dfdc} are used for cross-dataset evaluation, while DF40~\cite{df40} is used for cross-method evaluation following~\cite{x2dfd,agent4faceforgery}. We report AUC as the primary detection metric at the frame level, with video-level AUC additionally reported for the cross-dataset protocol by averaging frame scores within each video.

\begin{table*}[t]
    \centering
    \caption{\textbf{Frame-level} cross-method AUC (\%) on the six-subset DF40. The best results are highlighted in bold. All experimental results are taken from the original paper and~\cite{agent4faceforgery}.}
    \label{tab:df40_six_frame}
    \begin{tabular}{lcccccccc}
        \toprule
        Method & Venue & UniFace & E4S & FaceDancer & FSGAN
        & InSwap & SimSwap & Avg. \\
        \midrule
        RECCE~\cite{recce}              & CVPR'22   & 84.2 & 65.2 & 78.3 & 88.4 & 79.5 & 73.0 & 78.1 \\
        SBI~\cite{sbi}                & CVPR'22   & 64.4 & 69.0 & 44.7 & 87.9 & 63.3 & 56.8 & 64.4 \\
        CORE~\cite{core}               & CVPRW'22  & 81.7 & 63.4 & 71.7 & 91.1 & 79.4 & 69.3 & 76.1 \\
        IID~\cite{iid}                & CVPR'23   & 79.5 & 71.0 & 79.0 & 86.4 & 74.4 & 64.0 & 75.7 \\
        UCF~\cite{ucf}                & ICCV'23   & 78.7 & 69.2 & 80.0 & 88.1 & 76.8 & 64.9 & 76.3 \\
        LSDA~\cite{lsda}               & CVPR'24   & 85.4 & 68.4 & 75.9 & 83.2 & 81.0 & 72.7 & 77.8 \\
        CDFA~\cite{cdfa}               & ECCV'24   & 76.5 & 67.4 & 75.4 & 84.8 & 72.0 & 76.1 & 75.4 \\
        ProDet~\cite{prodet}             & NeurIPS'24& 84.5 & 71.0 & 73.6 & 86.5 & 78.8 & 77.8 & 78.7 \\
        X$^{2}$DFD-7B~\cite{x2dfd}         & NeurIPS'25& 85.5 & 90.3 & 82.6 & 91.1 & 81.2 & 85.1 & 86.0 \\
        FIA-USA~\cite{fia-usa}            & NeurIPS'25& 91.8 & 87.5 & 83.0 & 86.3 & 87.4 & 91.0 & 87.8 \\
        ResProto-FD~\cite{resproto}        & AAAI'26   & 90.3 & 94.8 & 87.8 & 94.9 & 89.3 & 85.7 & 90.5 \\
        Agent4FF~\cite{agent4faceforgery}  & CVPR'26   & 96.3 & 92.4 & 92.9 & 94.8 & 92.4 & \textbf{94.6} & 93.9 \\
        SRI-Net~\cite{sri-net}            & ICLR'26   & 92.0 & 89.4 & 95.3 & 94.2 & 91.1 & 83.3 & 90.9 \\
        \midrule
        Ours (Stage 1)     & --        & \textbf{97.2} & \textbf{98.7}
                           & \textbf{95.9} & \textbf{96.1}
                           & \textbf{94.3} & 94.0 & \textbf{96.0} \\
        Ours (Stage 2)     & --        & 95.6 & 98.6 & 94.3 & 95.3 & 90.6 & 91.9 & 94.4 \\
        \bottomrule
    \end{tabular}
\end{table*}

\begin{table*}[t]
\centering
\caption{Automatic and human evaluation of explanation quality. ``Corr.'' denotes the average MLLM-judge score over correctly classified samples, while ``All'' assigns a score of 1 to incorrect predictions following VERITAS~\cite{veritas}. Human Rank is obtained from a blinded ranking study on 100 samples drawn from the fixed 1K evaluation pool, where lower is better.}
\label{tab:opensource_judge}
\begin{tabular}{lccccccccc|c}
\toprule
\multirow{2}{*}{Method}
& \multicolumn{2}{c}{LLaVA-1.5-7B}
& \multicolumn{2}{c}{Qwen2.5-VL-7B}
& \multicolumn{2}{c}{InternVL2.5-8B}
& \multicolumn{2}{c}{Average}
& \multirow{2}{*}{ACC}
& \multirow{2}{*}{Human Rank $\downarrow$} \\
\cmidrule(lr){2-3}
\cmidrule(lr){4-5}
\cmidrule(lr){6-7}
\cmidrule(lr){8-9}
& Corr. & All
& Corr. & All
& Corr. & All
& Corr. & All
& \\
\midrule
LLaVA-1.5-7B
& 1.677 & 1.212
& 3.698 & 1.716
& 4.032 & 1.945
& 3.136 & 1.624
& \underline{0.313}
& 3.1313 \\

Qwen2.5-VL-7B
& \underline{2.996} & \underline{1.572}
& \textbf{3.996} & \underline{1.811}
& \textbf{4.661} & \underline{2.114}
& \underline{3.885} & \underline{1.832}
& 0.305
& \underline{2.5291} \\

InternVL2.5-8B
& 2.228 & 1.379
& 2.924 & 1.531
& 3.355 & 1.727
& 2.836 & 1.546
& \underline{0.313}
& 2.5992 \\

EviStress (Ours)
& \textbf{3.505} & \textbf{2.751}
& \underline{3.939} & \textbf{2.479}
& \underline{4.573} & \textbf{3.520}
& \textbf{4.006} & \textbf{2.917}
& \textbf{0.710}
& \textbf{1.7404} \\
\bottomrule
\end{tabular}
\end{table*}

\subsection{Implementation Details}
\label{subsec:detail}

We use CLIP ViT-L/14~\cite{radford2021learning} as the Stage-1 backbone with a frozen text encoder and LoRA-adapted visual encoder. LoRA is applied to the $q/k/v$ projections of all visual Transformer blocks with rank 6, $\alpha=12$, and dropout 0.8. Stage 1 is trained with a learning rate of $5\times10^{-6}$ and a batch size of 64. Unless otherwise specified, we set the perturbation depth to $K=1$, the Gaussian perturbation strength to $\sigma=0.04$, and the DVP weight to $\lambda_{\mathrm{dvp}}=0.2$. ETPP and DVP are used only during training and introduce no additional module at inference.

In Stage 2, the adapted CLIP encoder replaces the original LLaVA-v1.5-7B vision tower~\cite{llava} and remains frozen. We follow the image--text data and training protocol of X$^{2}$DFD~\cite{x2dfd}: the model is trained for three epochs with batch size 4 and warmup ratio 0.03; the multimodal projector uses a learning rate of $2\times10^{-5}$, while the remaining trainable parameters use $2\times10^{-4}$. LoRA on the language model uses rank 16 and $\alpha=32$. Only the multimodal projector and language-model LoRA parameters are optimized.

\subsection{Generalizability Evaluation}
\label{sec:generalization}

Tables~\ref{tab:traditional_frame}--\ref{tab:df40_six_frame} compare EviStress with recent detectors under cross-dataset and cross-method protocols. Stage 1 achieves an average frame-level AUC of 90.0\% and video-level AUC of 94.2\% on CDF, DFD, and DFDC, while reaching 96.0\% frame-level AUC on the six DF40 manipulation subsets. Stage 2 retains strong generalization after multimodal instruction tuning.

Stage 2 moderately decreases from Stage 1 as forgery-sensitive representations are mapped into multimodal space and jointly optimized with text generation. This alters fine-grained forensic evidence use, while explanations are not explicitly constrained to align with patch-level evidence. Stage 2 still generalizes well and enables joint prediction and explanation.

\subsection{Explanation Quality Evaluation}

Table~\ref{tab:opensource_judge} evaluates explanation quality using three open-source MLLMs as independent judges under the VERITAS~\cite{veritas} protocol. On correctly classified samples, EviStress achieves the highest average score of 4.006, compared with 3.885 for the strongest open-source baseline. When all samples are considered, its score remains substantially higher (2.917 vs.\ 1.832), reflecting both stronger detection accuracy and explanation quality.

\begin{figure*}[ht]
    \centering
    \includegraphics[width=1\linewidth]{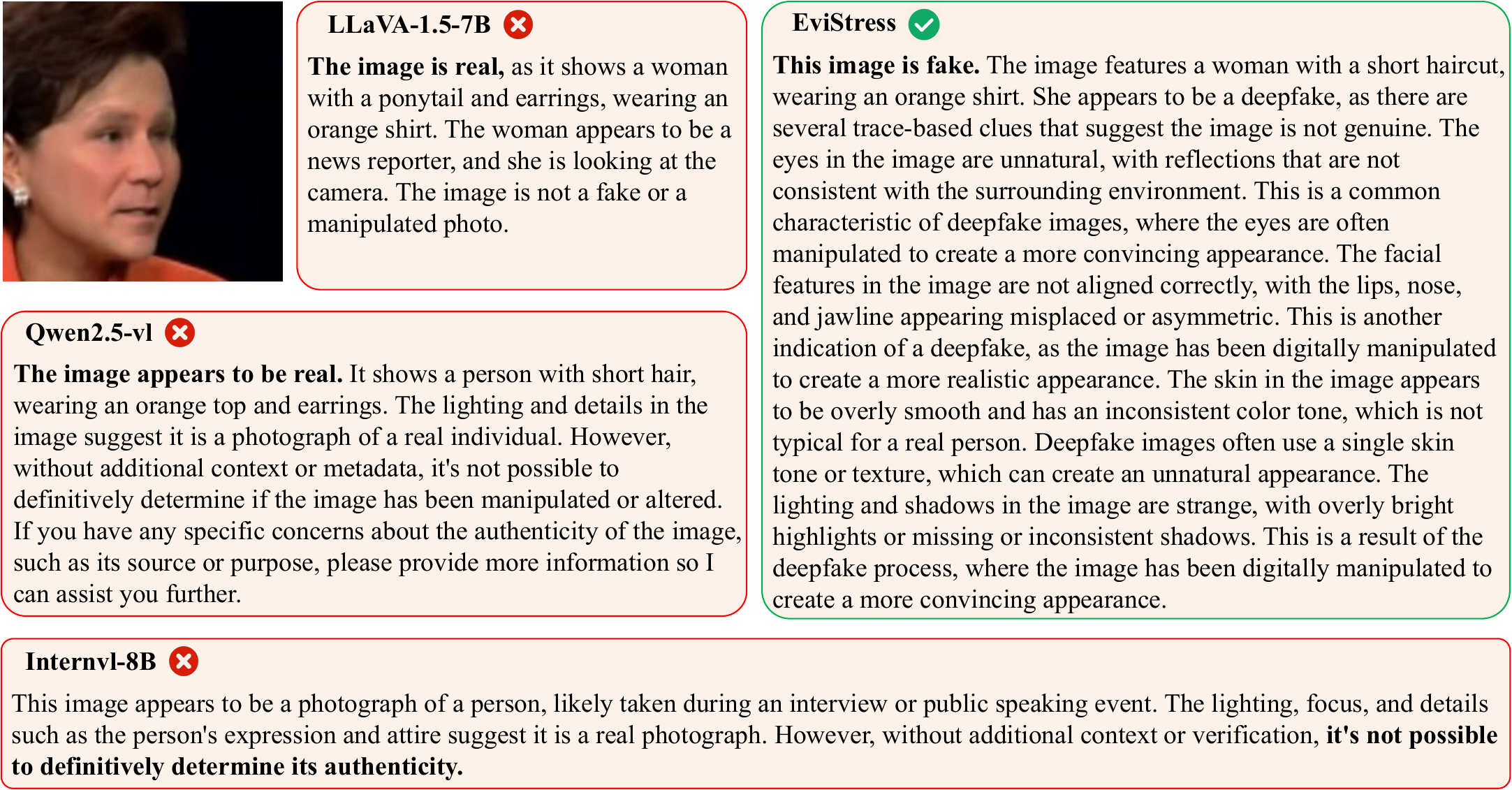}
    \caption{Fake image from the CDF dataset. All models use the same prompt text.}
    \label{fig:annotation_cdf}
    \Description{This image contains the original forged image and the detection explanations for that image provided by different models.}
\end{figure*}

We further conduct a human ranking study by randomly sampling 100 examples from the same fixed 1K evaluation pool. For each sample, the anonymous outputs of the four compared models are ranked according to overall explanation quality, and the model identities and presentation order are hidden during evaluation. EviStress obtains the best mean rank of 1.740, consistently outperforming the three open-source MLLMs. Figure~\ref{fig:annotation_cdf} provides a representative qualitative comparison.


\begin{table*}[t]
    \centering
    \caption{Robustness evaluation on CDF, DFD, and DFDC. For each corruption, AUC (\%) is averaged over five severity levels, and $\Delta$AUC denotes the drop from clean AUC ($\mathrm{AUC}_{clean}-\mathrm{AUC}_{corr}$). Avg. averages over the five corruption types. Best results within each dataset are highlighted in bold.}
    \label{tab:robustness_summary}
    \begin{tabular}{llc*{6}{cc}}
        \toprule
        & & & \multicolumn{2}{c}{Gaussian Noise} & \multicolumn{2}{c}{Block-wise} & \multicolumn{2}{c}{Color Contrast} & \multicolumn{2}{c}{Color Saturation} & \multicolumn{2}{c}{JPEG Compression} & \multicolumn{2}{c}{Average} \\
        Dataset & Model & Clean & AUC & $\Delta$ & AUC & $\Delta$ & AUC & $\Delta$ & AUC & $\Delta$ & AUC & $\Delta$ & AUC & $\Delta$ \\
        \midrule
        \multirow{2}{*}{CDF} & LoRA & 88.44 & 64.32 & 24.12 & 84.02 & 4.42 & 88.83 & -0.39 & 82.43 & 6.01 & 78.07 & 10.37 & 79.53 & 8.91 \\
         & ETPP+DVP & \textbf{91.35} & \textbf{70.63} & \textbf{20.72} & \textbf{89.11} & \textbf{2.24} & \textbf{91.50} & \textbf{-0.15} & \textbf{86.40} & \textbf{4.95} & \textbf{82.04} & \textbf{9.31} & \textbf{83.94} & \textbf{7.41} \\
        \midrule
        \multirow{2}{*}{DFD} & LoRA & \textbf{94.08} & 66.81 & 27.27 & 91.05 & 3.03 & \textbf{92.44} & \textbf{1.64} & \textbf{91.52} & 2.56 & 79.91 & 14.18 & 84.35 & 9.74 \\
         & ETPP+DVP & 93.67 & \textbf{67.63} & \textbf{26.04} & \textbf{91.08} & \textbf{2.59} & 91.82 & 1.85 & 91.44 & \textbf{2.23} & \textbf{80.04} & \textbf{13.63} & \textbf{84.40} & \textbf{9.27} \\
        \midrule
        \multirow{2}{*}{DFDC} & LoRA & 84.75 & 65.90 & 18.85 & 80.21 & 4.54 & 82.51 & \textbf{2.24} & 80.33 & 4.42 & \textbf{77.21} & \textbf{7.54} & 77.23 & 7.52 \\
         & ETPP+DVP & \textbf{85.11} & \textbf{68.43} & \textbf{16.68} & \textbf{81.49} & \textbf{3.62} & \textbf{82.68} & 2.43 & \textbf{81.13} & \textbf{3.98} & 77.11 & 8.00 & \textbf{78.17} & \textbf{6.94} \\
        \bottomrule
    \end{tabular}%
\end{table*}

\subsection{Robustness to Controlled Corruptions}
\label{subsec:robustness}
Table~\ref{tab:robustness_summary} evaluates the visual detector under five corruption types and five severity levels, using corruption tests based on~\cite{deeperforensics}. Compared with LoRA, ETPP+DVP improves the mean corrupted AUC on all three datasets and reduces the average AUC degradation from 8.91 to 7.41 points on CDF, from 9.74 to 9.27 on DFD, and from 7.52 to 6.94 on DFDC. The largest improvements occur under Gaussian noise and block-wise corruption, while the overall trend across heterogeneous perturbations confirms that targeted evidence stress improves robustness beyond a specific corruption type.

\subsection{Ablation Study}
\label{sec:abla}

\subsubsection{Perturbation strategy.}
Table~\ref{tab:probability_correlation} separates the effects of generic feature perturbation, dependence-aware allocation, and DVP. A fixed probability of 0.5 already improves over LoRA, showing that controlled feature perturbation provides useful regularization. Positive correlation achieves the best average AUC, whereas negative correlation underperforms the fixed-probability setting, demonstrating that the perturbation budget is more effective when stronger intervention is assigned to highly dependent patches. Removing $\mathcal{L}_{\mathrm{dvp}}$ slightly reduces the average performance, confirming its complementary role in preserving discriminative information within the induced variation.

\subsubsection{Influence of Patch-dependence cues.}
Table~\ref{tab:cue_ablation} evaluates the global (G), relational (R), and semantic (S) cues used for patch-dependence estimation. Relational and semantic cues are individually stronger than the global cue, and their combination provides the best two-cue result. Using all three cues further achieves the highest average AUC of 90.83\%, indicating that they provide complementary information for estimating patch dependence.

\begin{table}[t]
    \centering
    \caption{Ablation on the correlation between patch dependence and perturbation probability. All results are reported in frame-level AUC (\%). The best and second-best results are highlighted in bold and underlined, respectively.}
    \label{tab:probability_correlation}
    \begin{tabular}{l|cccc}
        \toprule
        Strategy & CDF & DFDC & DF40 & Avg. \\
        \midrule
        LoRA
        & 88.44
        & 84.75
        & 95.32
        & 89.50 \\
        $Prob=0.5$
        & 90.75
        & 84.91
        & \underline{95.74}
        & 90.47 \\
        Negative
        & 89.15
        & 85.10
        & 95.19
        & 89.81 \\
        w/o $L_{dvp}$
        & \underline{91.00}
        & \textbf{85.40}
        & 95.67
        & \underline{90.69} \\
        Positive
        & \textbf{91.35}
        & \underline{85.11}
        & \textbf{96.02}
        & \textbf{90.83} \\
        \bottomrule
    \end{tabular}
\end{table}

\begin{table}[t]
    \centering
    \caption{Ablation study of the three evidence cues. The experimental setup is consistent with Table~\ref{tab:probability_correlation}.}
    \label{tab:cue_ablation}
    \begin{tabular}{ccc cccc}
        \toprule
        \multicolumn{3}{c}{Evidence Cue} &
        \multicolumn{4}{c}{Dataset} \\
        \cmidrule(lr){1-3}
        \cmidrule(lr){4-7}
        G & R & S &
        CDF & DFDC & DF40 & Avg. \\
        \midrule
        $\checkmark$ & $\times$     & $\times$
        & 90.98 & 85.07 & 95.91 & 90.65 \\

        $\times$     & $\checkmark$ & $\times$
        & 91.12 & \textbf{85.21} & 95.92 & 90.75 \\

        $\times$     & $\times$     & $\checkmark$
        & 91.20 & 85.09 & \underline{95.97} & 90.75 \\

        $\checkmark$ & $\checkmark$ & $\times$
        & 91.09 & 85.08 & 95.94 & 90.70 \\

        $\checkmark$ & $\times$     & $\checkmark$
        & 91.24 & 85.11 & 95.94 & 90.76 \\

        $\times$     & $\checkmark$ & $\checkmark$
        & \underline{91.34} & \underline{85.16} & 95.91 & \underline{90.80} \\

        \midrule
        $\checkmark$ & $\checkmark$ & $\checkmark$
        & \textbf{91.35} & 85.11 &
        \textbf{96.02} & \textbf{90.83} \\
        \bottomrule
    \end{tabular}
\end{table}

\subsubsection{Perturbation depth.}
Table~\ref{tab:perturbation_depth} shows that applying ETPP only to the first Transformer block ($K=1$) achieves the best average AUC. Performance gradually decreases as the intervention is extended to deeper blocks, suggesting that evidence stress is most effective before local patch features are extensively aggregated into higher-level contextual representations. We therefore use $K=1$ in all main experiments.

\begin{table}[t]
    \centering
    \caption{Effect of perturbation depth. The experimental setup is consistent with Table~\ref{tab:probability_correlation}.}
    \label{tab:perturbation_depth}
    \begin{tabular}{c|cccc}
        \toprule
        $K$ & CDF & DFDC & DF40 & Avg. \\
        \midrule
        1  & \textbf{91.35} & \textbf{85.11}
           & \textbf{96.02} & \textbf{90.83} \\
        2  & \underline{91.04} & \underline{85.10}
           & 95.91 & \underline{90.68} \\
        3  & 90.96 & 84.94 & 95.83 & 90.58 \\
        4  & 90.86 & 84.51 & 95.91 & 90.43 \\
        8  & 90.83 & 84.51 & 95.86 & 90.40 \\
        10 & 90.72 & 84.54 & \underline{95.92} & 90.39 \\
        \bottomrule
    \end{tabular}
\end{table}

\begin{table}[t]
    \centering
    \caption{Sensitivity to the standard deviation $\sigma$ of the multiplicative Gaussian perturbation. The experimental setup is consistent with Table~\ref{tab:probability_correlation}.}
    \label{tab:perturbation_sigma}
    \begin{tabular}{c|cccc}
        \toprule
        $\sigma$
        & CDF & DFDC & DF40 & Avg. \\
        \midrule
        0.02
        & 89.64 & 84.88 & 95.55 & 90.02 \\
        0.03
        & 91.22 & 85.09 & \underline{95.97} & 90.76 \\
        0.04
        & \textbf{91.35}
        & \underline{85.11}
        & \textbf{96.02}
        & \textbf{90.83} \\
        0.05
        & \underline{91.23}
        & \textbf{85.19}
        & 95.94
        & \underline{90.79} \\
        0.06
        & 86.65 & 82.11 & 91.86 & 86.87 \\
        0.10
        & 90.70 & 84.81 & 95.89 & 90.47 \\
        0.15
        & 88.75 & 83.69 & 95.17 & 89.20 \\
        \bottomrule
    \end{tabular}
\end{table}

\begin{figure}[t]
    \centering
    \includegraphics[width=0.7\linewidth]{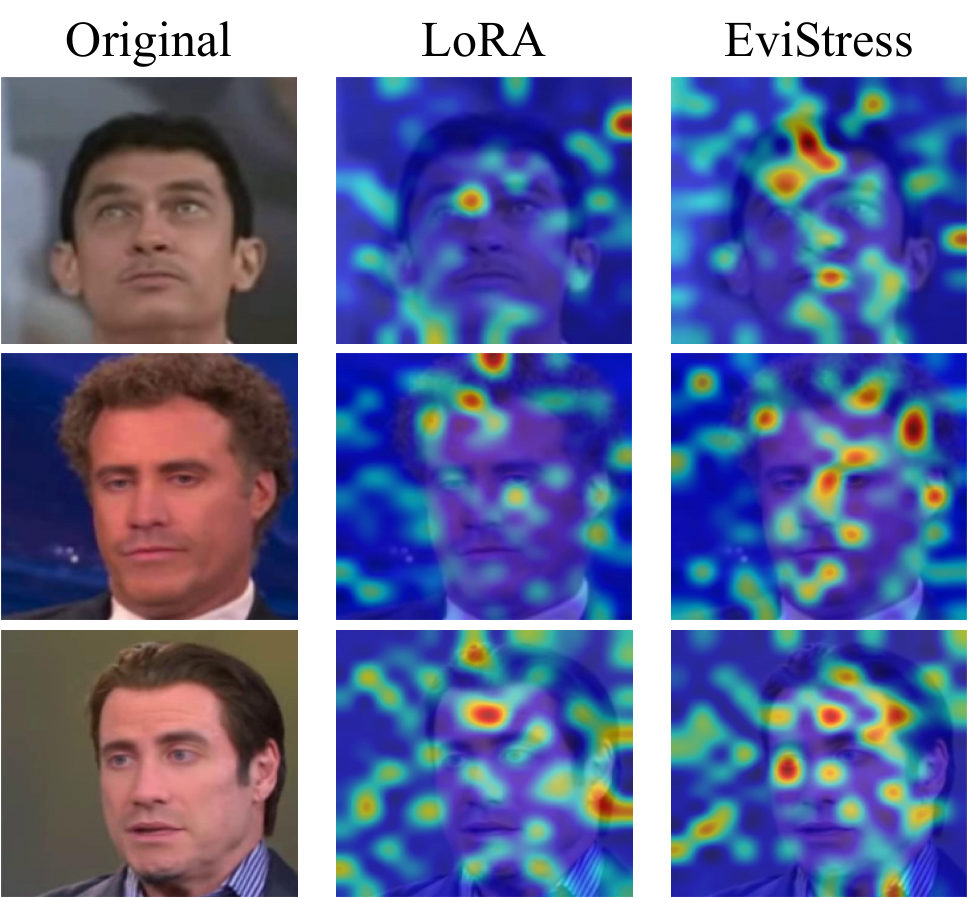}
    \caption{Grad-CAM visualization of LoRA and EviStress on representative unseen samples. EviStress exhibits visibly different response patterns across multiple facial regions compared with the more localized responses of LoRA.}
    \label{fig:visualization}
    \Description{Original face images and their Grad-CAM visualizations produced by standard LoRA and EviStress.}
\end{figure}

\subsubsection{Perturbation strength.}
Table~\ref{tab:perturbation_sigma} studies the Gaussian perturbation strength $\sigma$. Moderate values between 0.03 and 0.05 provide stable performance, with $\sigma=0.04$ achieving the highest average AUC. Weaker perturbation provides insufficient evidence stress, whereas excessive perturbation can distort useful discriminative information; hence $\sigma=0.04$ is used by default.

\subsubsection{Qualitative visualization.}
Figure~\ref{fig:visualization} compares the Grad-CAM responses of standard LoRA and EviStress on representative unseen samples. LoRA often exhibits strong responses in a few localized or peripheral regions, whereas EviStress redistributes its responses toward multiple facial areas associated with the authenticity decision. These qualitative differences complement the robustness results by illustrating distinct spatial response patterns between the two detectors.

\section{Conclusion}
\label{sec:conclusion}

We presented \textbf{EviStress}, a two-stage framework for generalizable face forgery detection with textual explanations, motivated by the authenticity challenges of online facial content. In Stage 1, \textbf{ETPP} preferentially perturbs highly dependent visual evidence, while \textbf{DVP} preserves authenticity discriminative information in the induced variation. The resulting encoder achieves strong cross-dataset and cross-method generalization and reduced average AUC degradation under the evaluated visual corruptions. In Stage 2, \textbf{IEIT} transfers the adapted encoder into an MLLM for joint authenticity prediction and textual explanation without relying on an external detector. A limitation is that generated rationales are not explicitly grounded in patch-level decision evidence, so explanation quality does not necessarily imply causal faithfulness. Future work will explore region--text alignment and intervention-based supervision to strengthen this connection.


\bibliographystyle{ACM-Reference-Format}
\bibliography{reference}

\appendix

\end{document}